\pdfoutput=1

\documentclass[11pt]{article}

\usepackage[final]{acl}

\usepackage{times}
\usepackage{latexsym}

\usepackage[T1]{fontenc}

\usepackage[utf8]{inputenc}

\usepackage{microtype}

\usepackage{inconsolata}

\usepackage{graphicx}

\usepackage{amsmath}
\usepackage{hyperref}

\title{Exploring OCR-augmented Generation for Bilingual VQA}

\author{
 \textbf{JoonHo Lee\thanks{\texttt{Work done while at KL-Net}}},
 \textbf{Sunho Park}
\\
\\
 KL-Net, South Korea
}

\begin{document}
\maketitle
\begin{abstract}
We investigate OCR-augmented generation with Vision Language Models (VLMs), exploring tasks in Korean and English toward multilingualism. To support research in this domain, we train and release KLOCR, a strong bilingual OCR baseline trained on 100M instances to augment VLMs with OCR ability. To complement existing VQA benchmarks, we curate KOCRBench for Korean VQA, and analyze different prompting methods. Extensive experiments show that OCR-extracted text significantly boosts performance across open source and commercial models. Our work offers new insights into OCR-augmented generation for bilingual VQA. Model, code, and data are available at \href{https://github.com/JHLee0513/KLOCR}{https://github.com/JHLee0513/KLOCR}.
\end{abstract}

\section{Introduction}

Optical Character Recognition (OCR) interprets text from visual inputs for applications such as accessibility, business automation, and robotics. The task requires understanding the spatial layout, semantic content, and inter-component relationships of text~\cite{nacson2024docvlmmakevlmefficient,wang-etal-2024-docllm}. Despite the progress, traditional OCR pipelines based on text detection and recognition exhibit limitations in scalability and human-level understanding~\cite{wei2024general}.

In this work, we explore the limits of OCR-augmented generation with Vision Language Models (VLMs). Recent advancements in VLMs show competitive OCR performance to traditional pipelines, and their semantic knowledge offers promising avenues toward end-to-end, OCR-capable agents.~\cite{9423358,masry-etal-2022-chartqa,Liu_2024, tang2024mtvqa, thomas-etal-2024-leveraging,Liu_2024} In comparison to OCR-free document understanding models~\cite{kim2022donut, blecher2023nougat}, VLMs are also capable of using their conversational abilities to directly address the downstream task at hand.

\begin{figure}[t]
    \centering
    \includegraphics[width=0.48\textwidth]{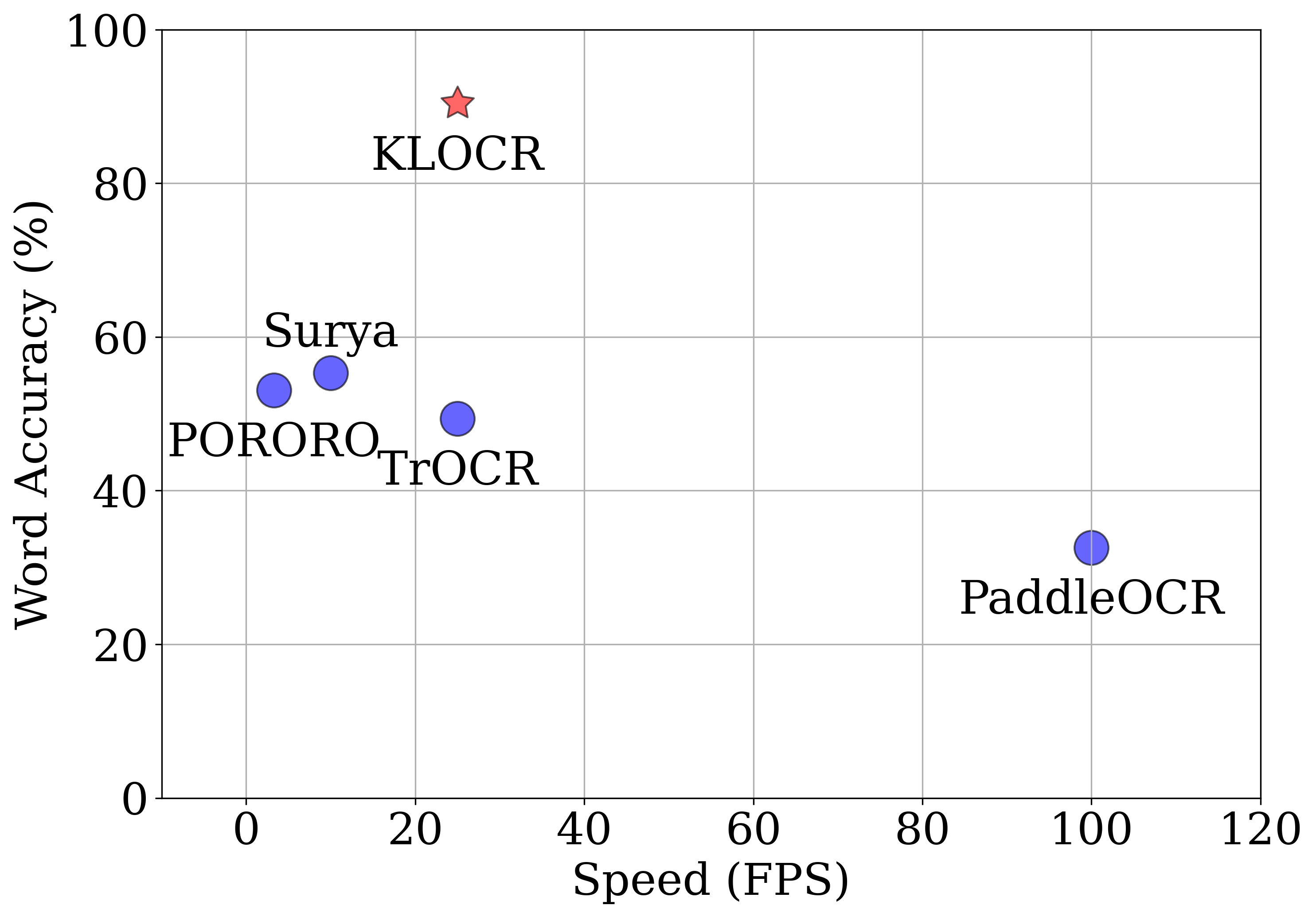}
     \caption{OCR model comparison on the validation set of KLOCR data. KLOCR not only sets state-of-the-art accuracy on the benchmark, but also exhibits the best accuracy-speed tradeoff.}
    \label{fig:klocr-comparison}
\end{figure}

We investigate OCR-augmented generation for Visual Question and Answering in English and Korean, with aims to promote research of multilingual models. We provide KOCRBench, a novel Korean OCR Benchmark, and KLOCR, a robust bilingual OCR baseline. Our contribution lies in exploring the impact of OCR in providing additional context to VLMs, and we anticipate the benchmark and OCR model will encourage further research. Extensive experiments show that OCR significantly boosts performance, indicating room for further improvement by VLMs. Overall, findings show the presence of character-accurate key information was the most crucial factor to model success. Model and code are available at \href{https://github.com/JHLee0513/KLOCR}{https://github.com/JHLee0513/KLOCR}.

\section{Related Work}
\subsection{Text Recognition}

Text recognition~\cite{7801919,li2021trocr,du2022svtrscenetextrecognition,Rang_2024_CVPR,zhao2024clip4str} forms the core algorithm behind OCR. \citet{Rang_2024_CVPR} demonstrated scaling laws present in OCR with common English benchmarks~\cite{10.1109/ICCV.2011.6126402,iii5k,ic13,svtp,cute,ic15}. We follow this insight to collect large-scale training data for KLOCR.

\subsection{Scene Text Detection}
Scene Text Detection~\cite{DBLP:journals/corr/abs-1904-01941,liao2020real,ye2022dptextdetrbetterscenetext,liao2022real,ye2022dptext} identifies text regions as bounding boxes, assisting recognition and improving spatial understanding. We integrate KLOCR with PaddleOCR~\cite{DBLP:journals/corr/abs-2009-09941,li2022ppocrv3attemptsimprovementultra} implementation of DBNet~\cite{liao2022real} for our experiments.

\subsection{Document Structure Analysis}
Document Structure Analysis enhances OCR by identifying the structure of the text such as reading order, text types, and layout. Prior work includes structure analysis~\cite{doclaynet2022, Da_2023_ICCV}, table detection and recognition~\cite{smock2022pubtables,peng2023high,peng2024unitable,peng2024self}, reading order detection~\cite{wang2021layoutreaderpretrainingtextlayout}, and semantic structure analysis~\cite{8099945}. Despite their strong in-domain accuracy, the models require a significant amount of densely annotated data and show limited performance for out-of-domain samples~\cite{zhong2019publaynetlargestdatasetdocument}.

\subsection{Key Information Extraction}

Key Information Extraction (KIE)~\cite{vies2021,10203796} 
focuses on extracting queried information rather than converting the entire visual input to text. Public benchmarks such as FUNSD~\cite{jaume2019} and SROIE~\cite{Huang_2019} verify the extraction capabilities of pipelines and models in receipts, records, and other documents. As many applications rely on this task, we include it in KOCRBench.

\subsection{Vision Language Models}

Vision Language Models are general-purpose models trained on large amounts of image and text data for conversational vision language tasks~\cite{Qwen2.5-VL, alayrac2022flamingovisuallanguagemodel,Li2022BLIPBL,geminiteam2024geminifamilyhighlycapable,NEURIPS2023_6dcf277e,li2022blip,nvlm2024}. Their recent applications in vision language tasks and even embodied AI demonstrate their wide range of capabilities~\cite{rt22023arxiv,kim24openvla}.

\section{KLOCR: Open Source Bilingual OCR Model}

\citet{Rang_2024_CVPR} demonstrated scaling laws in OCR, achieving state-of-the-art performances on six common English benchmarks by training a transformer based model on a large-scale dataset. Following this insight, we train the Korean Language Optical Character Recognition (KLOCR) model on a 100M\footnote{Total dataset size is +120M, while we hold out $\sim$20M as validation.} instances bilingual dataset, achieving competitive performance on English and state-of-the-art accuracy on Korean. 

\subsection{Data}

\begin{figure}[t]
    \centering
    \includegraphics[width=0.45\textwidth]{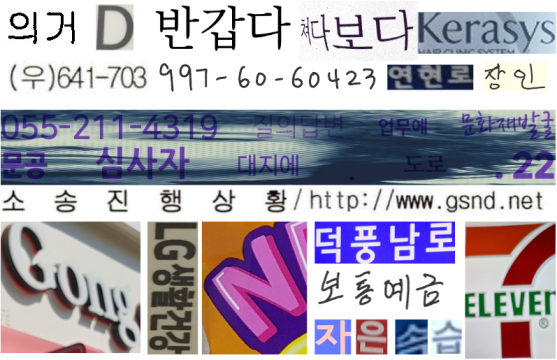}
    \caption{Samples from KLOCR data mixture. The data collection is bilingual and varies across multiple domains (e.g. documents, road signs, handwriting).}
    \label{fig:klocr-viz}
\end{figure}

We curate a diverse mixture of English and Korean OCR data, varying in text length and image domain. Table~\ref{tab:kl-data} describes our final composition, where most of the data is sourced from multilingual datasets made publicly available at AI-Hub. We combine SynthTIGER-v1.1~\cite{yim2021synthtiger}, PixParse ~\cite{pixparse_idl_wds}, and generate 3M samples of multi-line, multi-word samples to increase data variety. We split the final collection into approximate 80-20 split for training and testing. Figure~\ref{fig:klocr-viz} highlights several samples that can be found in our mixture. We share the AIHub dataset details, licensing information, and pre-processing steps taken in Appendix~\ref{sec:app-data}.

\begin{table}[ht]
\centering
\begin{small}
\begin{tabular}{cccc}
\hline
\textbf{Type} & \textbf{Lang} & \textbf{Dataset} & \textbf{Instances} \\ \hline
Real & Ko+En        & AIHub    & 100M   \\ \hline
Real & En    & PixParse  & 7.2M  \\ \hline
Real  & En     & Union14M    & 3.2M   \\ \hline
Synth & En    & SynthTIGER  & 10M  \\ \hline
Synth  & Ko+En  & SynthTIGER\(\dagger\)  & 3M  \\ \hline
Real  & En   & UberText  & 0.1M  \\ \hline
Real  & En   & TextOcr  & 0.7M  \\ \hline
Real  & En   & CocoText  & 0.07M  \\ \hline
Mixed  & Ko+En   & \textbf{Total}  & 124.3M  \\ \hline
\end{tabular}
\end{small}
\caption{KLOCR Data Mixture. \(\dagger\)We generate additional data by running SynthTIGER data engine with the text from the AIHub datasets. After validation split, we have approximately 100M training samples.}
\label{tab:kl-data}
\end{table}

\subsection{Model}

We finetune TrOCR~\cite{li2021trocr} pretrained on a custom synthetic dataset generated with the SynthTIGER engine~\cite{team-lucid_trocr_small_korean}. The model uses DeiT~\cite{pmlr-v139-touvron21a} as its encoder and RoberTa~\cite{DBLP:journals/corr/abs-1907-11692} as its decoder. 
At 55M Parameters, the model runs real-time (20+ FPS) on a desktop GPU. 

\subsection{Training}

We trained KLOCR for two epochs using two RTX A6000 GPUs, with a batch size of 64 per GPU. We use the AdamW~\cite{loshchilov2018decoupled} optimizer with a fixed learning rate of $5e^{-7}$ to avoid drifting too far from the initialized weights. Since most of the samples are clear high-quality images, we found data augmentation (random rotation, random brightness, CoarseDropOut~\cite{Devries2017ImprovedRO,Zhong_Zheng_Kang_Li_Yang_2020}) beneficial to model generalization.
The training run finishes in approximately 500 GPU hours, and we estimate the total development cost of the model to have been below 3000 GPU hours.

\section{Visual Question answering with OCR-Augmented Reasoning}
\label{sec:vqa}
We consider \textbf{Base} as a baseline method, where we prompt the VLM with the input image and query without any additional context. In comparison, OCR-based prompting, which we denote as \textbf{OCR}, prompts the VLM with OCR-extracted text as additional context.  We follow a format similar to \citet{NEURIPS2023_6dcf277e} but omit the bounding box coordinates. 

\begin{figure}[t]
    \centering
    \includegraphics[width=0.48\textwidth]{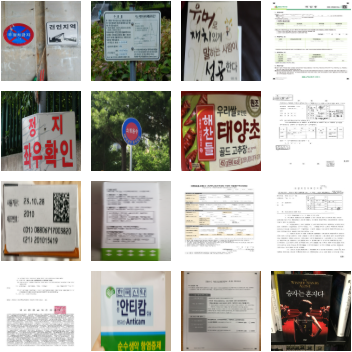}
    \caption{Sample images from the KOCRBench dataset. We collect various samples from KLOCR data mixture and repurpose samples from KVQA to create (image, question, answer) triplets. The dataset covers various scenarios with road signs, product images, and documents. Images have been resized for visualization purposes.}
    \label{fig:kocrbench-samples}
\end{figure}

\subsection{KOCRBench: Korean VQA Benchmark}

We curated KOCRBench to test VLMs' ability to handle visual question answering in Korean. Following design of the prior work in English OCR ~\cite{singh2019towards,9423358,masry2022chartqabenchmarkquestionanswering, Liu_2024}, we collected 250 questions from public sources spanning over 248 input images. Specifically, a portion of the benchmark is repurposed from the Korean Localization of Visual Question Answering for Blind People (KVQA)~\cite{kim2019korean} dataset with reinforced annotations. We generated the majority of samples by selecting raw images from the holdout data from KLOCR data mixture and annotated manually. We created annotations for 4 tasks: text recognition (22 samples), scene VQA (70 samples), document VQA (29 samples), and key information extraction (129 samples). The number of samples were based on our internal assessment of the importance of each task in real business processes.

\section{Experiments}

\begin{table}[htbp]
\centering
\begin{small}
\begin{tabular}{ccc}
\hline
\textbf{Method} & \textbf{CER\(\downarrow\)} &  \textbf{Word Accuracy\(\uparrow\)} \\ \hline
CLIP4STR-L*      & 125.2\%    &   9.0\%   \\
Surya         & 60.9\%  &   55.3\%    \\ 
PaddleOCR     & 49.6\%  &  32.6\%    \\ 
PORORO        & 30.0\%    &  53.1\%   \\
TrOCR         & 27.0\%  &  49.4\%    \\
\textbf{KLOCR}          &  \textbf{2.34\%}  & \textbf{94.6\%}  \\ \hline
\end{tabular}
\end{small}
\caption{\label{tab:accuracy_korean} Character Error Rate and word accuracy on the Korean OCR benchmark. KLOCR demonstrates significantly better performance than other open source models. \(\dagger\) denotes variant trained with additional Union14M-L dataset, matching its data distribution closer to the common English benchmarks. *Model from ~\citet{rang2024empiricalstudyscalinglaw} is only trained on English data, and therefore shows high error.}
\end{table}

\begin{table*}[htbp]
  \centering
  \begin{small}
  \begin{tabular}{@{}llccccccc@{}}
    \hline
    Method & IC13 & IIIT5k & SVT & CUTE80 & IC15 & SVTP & Avg \\
    \hline
    TrOCR                & 66.86 & 59.07 & 60.43 & 45.83 & 49.48 & 49.46 & 55.19 \\
    PORORO               & 78.30 & 64.30 & 56.57 & 47.57 & 45.33 & 46.05 & 56.35 \\
    Surya                & 82.73 & 71.50 & 74.19 & 44.79 & 64.00 & 64.19 & 69.48 \\
    \textit{KLOCR}       & \textit{95.92} & \textit{86.50} & \textit{93.20} & \textit{91.67} & \textit{84.87} & \textit{87.91} & 88.13 \\
    \hline
    CLIP4STR-L*  & 99.42 & 99.13 & 98.61 & 99.65 & 92.6  & 98.13 & 97.42 \\
    \hline
  \end{tabular}
  \end{small}
  \caption{\label{tab:word_accuracy_english}Word accuracy on English benchmarks. Avg is the total average accuracy across all samples from the benchmarks. CLIP4STR-L* trained by ~\citet{rang2024empiricalstudyscalinglaw} includes training splits of benchmark data in their training data. Despite not targeting the English benchmarks and using a much smaller model, KLOCR performance remains competitive.}
\end{table*}

\subsection{Implementation Details}

We used vLLM~\cite{kwon2023efficient} to host the VLMs on our hardware and hosted the OCR models on the same machine or on a separate machine with an RTX A1000 GPU. We conducted our experiments in PyTorch~\cite{paszke2019pytorchimperativestylehighperformance}.

\subsection{OCR Benchmarks}

Table~\ref{tab:accuracy_korean} provides evaluation on Korean OCR for currently available open source OCR models. KLOCR outperforms prior models by a significant margin, achieving 94.6\% word accuracy and 2.34\% character error rate. The performance gap between TrOCR and KLOCR despite the two sharing the same architecture highlights the importance of scaling up OCR data. As expected, the Clip4STR model by ~\citet{rang2024empiricalstudyscalinglaw} does not handle Korean and therefore achieves low accuracy. 

Table~\ref{tab:word_accuracy_english} provides evaluation on the six common English benchmarks. KLOCR demonstrates comparable performance without any in-domain training data, demonstrating its scale and variety. KLOCR significantly out-performs prior OCR models focusing on Korean. As a reference point, we include the CLIP4STR-L model trained by ~\citet{rang2024empiricalstudyscalinglaw}, which includes the training subset of the benchmark data in its training and evidently achieves the highest performance.

\subsection{Multilingual VQA}

\begin{table*}[bhtp]
\centering
\begin{small}
\begin{tabular}{ccccccc}
\hline
\textbf{Model} & \textbf{Prompt} &  \textbf{Recognition} &  \textbf{Scene} &  \textbf{Document} &  \textbf{KIE}  & \textbf{Total} \\ \hline
Qwen2.5-VL-7B        & Base    &  22  &  66  &  16  &  94   &  198  \\ 
Qwen2.5-VL-7B        & OCR   &  21  &  65  &  22   &  104   &  212  \\ 
\hline
InternVL2.5-7B        & Base    &  16  &  46  & 5  & 20   &  87  \\ 
InternVL2.5-7B        & OCR   &  19  &  52  &  10   &  81   &  162  \\ 
\hline
Qwen2.5-VL-32B-Instruct\(\dagger\) & Base    &  21  &  60  &  20  &  75   &  176  \\ 
Qwen2.5-VL-32B-Instruct\(\dagger\) & OCR    &  20  &  61  &  21  &  103   &  205  \\ 
\hline
gemini 2.0 flash        & Base    &  20  &  65  &  22  &  93   &  200  \\ 
gemini 2.0 flash        & OCR    &   19 &  64  &  23  &  97   &  203  \\ 
\hline
gemini-2.5-flash-preview-04-17        & Base    &  21  &  70  &  20  &  71   &  182  \\ 
gemini-2.5-flash-preview-04-17        & OCR    &  19  &  69  &  22 &  102   &  212  \\ 
\hline
\end{tabular}
\end{small}
\caption{KOCRBench Performance Comparison, for models with both base and instruction-tuned available, instruction-tuned variants are tested.\(\dagger\) Due to memory constraints, we run the AWQ quantized model.}
\label{tab:kocrbench_wide} 
\end{table*}

As aforementioned in Section ~\ref{sec:vqa}, we compare \textbf{Base} and \textbf{OCR} prompting. Table~\ref{tab:kocrbench_wide} shows the benchmark results across 5 models: Qwen-VL 2.5 7B, 32B~\cite{Qwen2.5-VL}, InternVL 2.5 7B~\cite{chen2024internvl}, Gemini 2.0 Flash, and Gemini 2.5 Flash~\cite{geminiteam2024geminifamilyhighlycapable}. The chosen models have shown competitive performances on the English benchmarks and also provide multilingual support. The Gemini models have been added to provide a reference point for commercially available models.

The addition of OCR-extracted information significantly improves accuracy for all models, aligning with the findings by \cite{rang2024empiricalstudyscalinglaw}. The largest improvements are observed from smaller models with a weaker base performance such as InternVL, indicating the OCR information is used by the models to correct their responses. Notably, we observe very strong base performance from Qwen-VL 2.5 7B despite its smaller size, indicating the potential fact that Qwen trainig mixture has substantial multilingual data.

Our results indicate largest performance improvement in Key Information Extraction, highlighting the usefulness of OCR's accurate character recognition. This also implies VLMs are yet to resolve spelling errors, especially on unusual and semantically meaningless words or obscure jargon.

\section{Discussion}
We further discuss the applicability of OCR-augmented generation with a set of ablation studies.
\textbf{When is OCR useful?} While KLOCR has shown robust performance and significantly boosted VLMs' performance in VQA, the trade-off between training OCR models and finetuning VLMs to improve their OCR ability should be weighed properly. Results on English~\cite{rang2024empiricalstudyscalinglaw} and Korean indicate OCR can play a crucial role in assisting VLMs, especially for low base performance models. It is also possible to finetune the VLMs directly on the OCR data, albeit with potential forgetting of other abilities. Meanwhile, it's challenging to train large-scale OCR model for low-resource languages, and hence resolving this issue for VLMs and OCR models remain a challenge.

\textbf{Impact of OCR accuracy on VLM} We verify the effectiveness of OCR-augmented generation by testing Qwen-VL 2.5 7B and InternVL 2.5 7B using KLOCR and TrOCR as the OCR extraction model. Results in Table~\ref{tab:ablation_ocr} clearly indicate that improvement in OCR also leads to an improvement in VLM response, while stronger models such as Qwen 2.5 show greater robustness against OCR error. 

\begin{table}[htbp]
  \centering
  \begin{small}
  \begin{tabular}{ccccccc}
    \hline
    VLM & OCR & \textbf{R} &  \textbf{S} &  \textbf{D} &  \textbf{K}  & \textbf{Total} \\
    \hline
    InternVL & TrOCR  & 18 &  54 & 8  &  47  & 127 \\
    InternVL & KLOCR  & 19 &  52 &  10 &  81  & 162 \\
    Qwen 2.5 & TrOCR  & 19 &  68 &  23 &  92  &  202 \\
    Qwen 2.5 & KLOCR  & 21 &  65 &  22 &  104  & 212 \\
    \hline
  \end{tabular}
  \end{small}
  \caption{\label{tab:ablation_ocr}Ablation study on OCR model. Using a more powerful OCR model (KLOCR) improves overall score.}
\end{table}

\begin{figure}[t]
    \centering
    \includegraphics[width=0.48\textwidth]{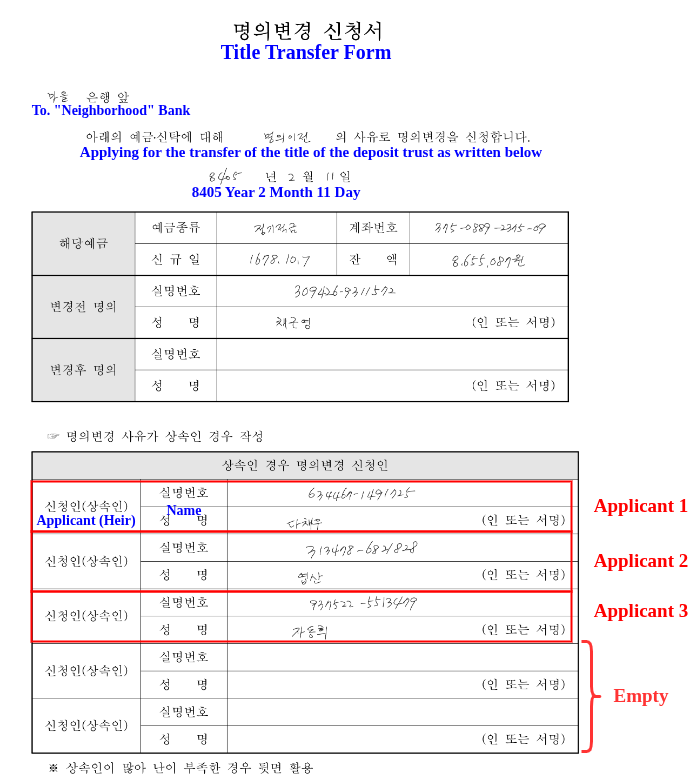}
    \caption{Example failure case of miscounting. Blue text indicates translated text for context. Boxed areas with red text highlight three applications written down. When asked to count the number of applicants in the form, VLMs often response to mistakenly list 5 valid applicants instead of 3.}
    \label{fig:vqoar-counting}
\end{figure}

\textbf{KOCRBench error analysis}
Our results on KOCRBench exhibit VLMs' weaknesses: 
\begin{enumerate}
    \item Counting: Counting has been a challenging task for either LLMs or VLMs~\cite{bigverdi2024perception}, and it is no exception in this case. As illustrated by the example in Figure~\ref{fig:vqoar-counting}, counting is a common source of error.
    \item Character-level precision: Observations show that misspelling and punctuation errors are the most common sources of error. While OCR-augmented generation generally alleviates this issue as observed in Table~\ref{tab:kocrbench_wide}, the approach may still struggle with edge cases.
    \item Refusing to answer: we observe several instances of refusal to answer where the VLM determines the question is unanswerable, with such cases more frequent with long context.
\end{enumerate}

\begin{table}[htbp]
  \centering
  \begin{small}
  \begin{tabular}{cccccc}
    \hline
    Gemini 2.5 & \textbf{R} &  \textbf{S} &  \textbf{D} &  \textbf{K}  & \textbf{Total} \\
    \hline
    Flash  & 19  &  69  &  22 &  102   &  212  \\
    Thinking  & 21 &  70 &  23 &  70\textcolor{blue}{(95)}  &  184\textcolor{blue}{(209)} \\
    \hline
  \end{tabular}
  \end{small}
  \caption{\label{tab:ablation_thinking}Ablation study on applying test-time scaling. Both methods are fed the OCR tokens as additional context. Scores in \textcolor{blue}{()} indicate what the model would have received if punctuation errors were not considered.}
\end{table}

\textbf{Does test-time scaling improve OCR-augmented generation?} We investigate whether test-time scaling~\cite{openai2024openaio1card,deepseekai2025deepseekr1incentivizingreasoningcapability,muennighoff2025s1simpletesttimescaling} improves OCR-augmented generation. Open source vision language models do not yet support reasoning in conjunction to vision at the time of our experiments, and therefore we run our experiments on \textit{gemini-2.5-flash-preview-04-17}, which supports reasoning with its "thinking" option. Results in Table~\ref{tab:ablation_thinking} indicate reasoning does not improve VQA capabilities, in particular due to significant a drop in KIE performance. Closer analysis showed that the model showed increased punctuation and spelling error with thinking, and often ignored OCR information more than the non-thinking variant. The punctuation errors in this case mostly are spacing errors specific to the Korean language. We manually check incorrect answers due to spacing errors in KIE, and observe that 25 errors were caused by this error. Had the score not account for this type of error, we would have observed a score of 209 that is much closer to the non-thinking variant. Therefore, our findings indicate reasoning models in multilingual VQA still holds more room for improvements. 

\section{Conclusion}

We introduced KOCRBench, a collection of text-oriented visual question and answering data for benchmarking Korean VQA towards multilingual visual understanding. Using the benchmark and our released KLOCR OCR model, we ran extensive experiments to explore the benefits and limitations of OCR-augmented generation for VQA. We observe that OCR most benefits the models by assisting them in precise character recognition. Our results indicate room for improving VLMs in more precise recognition and building an accurate representation of documents.

\section*{Limitations}

\textbf{KLOCR} While the 100M dataset is large-scale and publicly sourced, it relies heavily on AIHub and SynthTIGER. AIHub is only a data platform and the data sources are independent, but we expect more robustness if other sources could be used (e.g. the web), and if it can integrate more synthetic data and other large datasets e.g. REBU-Syn. Due to increasing scale and compute requirements, we leave this to future work. Additionally, as the focus of KLOCR is in its bilingual abilities, no tuning has been made to achieve state-of-the-art performance for English. Lastly, we leave expansion to other languages, especially low-resource ones, to future work.

\textbf{KOCRBench} KOCRBench captures various tasks in different domain scenarios, but its modest size of 250 questions does not fully capture the performance of models like other massive English VQA benchmarks. We aim to continue our work in curating data to expand the benchmark, and experiment with synthetic dataset creation to reduce the limitation of manual labeling. We anticipate our efforts to encourage other researchers to contribute to expanding multilingual VQA benchmarks. 

\bibliography{acl_latex}

\begin{thebibliography}{69}
\providecommand{\natexlab}[1]{#1}

\bibitem[{Alayrac et~al.(2022)Alayrac, Donahue, Luc, Miech, Barr, Hasson, Lenc, Mensch, Millican, Reynolds, Ring, Rutherford, Cabi, Han, Gong, Samangooei, Monteiro, Menick, Borgeaud, Brock, Nematzadeh, Sharifzadeh, Binkowski, Barreira, Vinyals, Zisserman, and Simonyan}]{alayrac2022flamingovisuallanguagemodel}
Jean-Baptiste Alayrac, Jeff Donahue, Pauline Luc, Antoine Miech, Iain Barr, Yana Hasson, Karel Lenc, Arthur Mensch, Katie Millican, Malcolm Reynolds, Roman Ring, Eliza Rutherford, Serkan Cabi, Tengda Han, Zhitao Gong, Sina Samangooei, Marianne Monteiro, Jacob Menick, Sebastian Borgeaud, and 8 others. 2022.
\newblock \href {https://arxiv.org/abs/2204.14198} {Flamingo: a visual language model for few-shot learning}.
\newblock \emph{Preprint}, arXiv:2204.14198.

\bibitem[{Baek et~al.(2019)Baek, Lee, Han, Yun, and Lee}]{DBLP:journals/corr/abs-1904-01941}
Youngmin Baek, Bado Lee, Dongyoon Han, Sangdoo Yun, and Hwalsuk Lee. 2019.
\newblock \href {https://arxiv.org/abs/1904.01941} {Character region awareness for text detection}.
\newblock \emph{CoRR}, abs/1904.01941.

\bibitem[{Bai et~al.(2025)Bai, Chen, Liu, Wang, Ge, Song, Dang, Wang, Wang, Tang, Zhong, Zhu, Yang, Li, Wan, Wang, Ding, Fu, Xu, Ye, Zhang, Xie, Cheng, Zhang, Yang, Xu, and Lin}]{Qwen2.5-VL}
Shuai Bai, Keqin Chen, Xuejing Liu, Jialin Wang, Wenbin Ge, Sibo Song, Kai Dang, Peng Wang, Shijie Wang, Jun Tang, Humen Zhong, Yuanzhi Zhu, Mingkun Yang, Zhaohai Li, Jianqiang Wan, Pengfei Wang, Wei Ding, Zheren Fu, Yiheng Xu, and 8 others. 2025.
\newblock Qwen2.5-vl technical report.
\newblock \emph{arXiv preprint arXiv:2502.13923}.

\bibitem[{Bigverdi et~al.(2024)Bigverdi, Luo, Hsieh, Shen, Chen, Shapiro, and Krishna}]{bigverdi2024perception}
Mahtab Bigverdi, Zelun Luo, Cheng-Yu Hsieh, Ethan Shen, Dongping Chen, Linda~G Shapiro, and Ranjay Krishna. 2024.
\newblock Perception tokens enhance visual reasoning in multimodal language models.
\newblock \emph{arXiv preprint arXiv:2412.03548}.

\bibitem[{Blecher et~al.(2023)Blecher, Cucurull, Scialom, and Stojnic}]{blecher2023nougat}
Lukas Blecher, Guillem Cucurull, Thomas Scialom, and Robert Stojnic. 2023.
\newblock \href {https://arxiv.org/abs/2308.13418} {Nougat: Neural optical understanding for academic documents}.
\newblock \emph{Preprint}, arXiv:2308.13418.

\bibitem[{Brohan et~al.(2023)Brohan, Brown, Carbajal, Chebotar, Chen, Choromanski, Ding, Driess, Dubey, Finn, Florence, Fu, Arenas, Gopalakrishnan, Han, Hausman, Herzog, Hsu, Ichter, Irpan, Joshi, Julian, Kalashnikov, Kuang, Leal, Lee, Lee, Levine, Lu, Michalewski, Mordatch, Pertsch, Rao, Reymann, Ryoo, Salazar, Sanketi, Sermanet, Singh, Singh, Soricut, Tran, Vanhoucke, Vuong, Wahid, Welker, Wohlhart, Wu, Xia, Xiao, Xu, Xu, Yu, and Zitkovich}]{rt22023arxiv}
Anthony Brohan, Noah Brown, Justice Carbajal, Yevgen Chebotar, Xi~Chen, Krzysztof Choromanski, Tianli Ding, Danny Driess, Avinava Dubey, Chelsea Finn, Pete Florence, Chuyuan Fu, Montse~Gonzalez Arenas, Keerthana Gopalakrishnan, Kehang Han, Karol Hausman, Alex Herzog, Jasmine Hsu, Brian Ichter, and 35 others. 2023.
\newblock Rt-2: Vision-language-action models transfer web knowledge to robotic control.
\newblock In \emph{arXiv preprint arXiv:2307.15818}.

\bibitem[{Chen et~al.(2024)Chen, Wu, Wang, Su, Chen, Xing, Zhong, Zhang, Zhu, Lu et~al.}]{chen2024internvl}
Zhe Chen, Jiannan Wu, Wenhai Wang, Weijie Su, Guo Chen, Sen Xing, Muyan Zhong, Qinglong Zhang, Xizhou Zhu, Lewei Lu, and 1 others. 2024.
\newblock Internvl: Scaling up vision foundation models and aligning for generic visual-linguistic tasks.
\newblock In \emph{Proceedings of the IEEE/CVF Conference on Computer Vision and Pattern Recognition}, pages 24185--24198.

\bibitem[{Da et~al.(2023)Da, Luo, Zheng, and Yao}]{Da_2023_ICCV}
Cheng Da, Chuwei Luo, Qi~Zheng, and Cong Yao. 2023.
\newblock Vision grid transformer for document layout analysis.
\newblock In \emph{Proceedings of the IEEE/CVF International Conference on Computer Vision (ICCV)}, pages 19462--19472.

\bibitem[{Dai et~al.(2024)Dai, Lee, Wang, Yang, Liu, Barker, Rintamaki, Shoeybi, Catanzaro, and Ping}]{nvlm2024}
Wenliang Dai, Nayeon Lee, Boxin Wang, Zhuolin Yang, Zihan Liu, Jon Barker, Tuomas Rintamaki, Mohammad Shoeybi, Bryan Catanzaro, and Wei Ping. 2024.
\newblock Nvlm: Open frontier-class multimodal llms.
\newblock \emph{arXiv preprint}.

\bibitem[{DeepSeek-AI(2025)}]{deepseekai2025deepseekr1incentivizingreasoningcapability}
DeepSeek-AI. 2025.
\newblock \href {https://arxiv.org/abs/2501.12948} {Deepseek-r1: Incentivizing reasoning capability in llms via reinforcement learning}.
\newblock \emph{Preprint}, arXiv:2501.12948.

\bibitem[{Devries and Taylor(2017)}]{Devries2017ImprovedRO}
Terrance Devries and Graham~W. Taylor. 2017.
\newblock \href {https://api.semanticscholar.org/CorpusID:23714201} {Improved regularization of convolutional neural networks with cutout}.
\newblock \emph{ArXiv}, abs/1708.04552.

\bibitem[{Du et~al.(2022)Du, Chen, Jia, Yin, Zheng, Li, Du, and Jiang}]{du2022svtrscenetextrecognition}
Yongkun Du, Zhineng Chen, Caiyan Jia, Xiaoting Yin, Tianlun Zheng, Chenxia Li, Yuning Du, and Yu-Gang Jiang. 2022.
\newblock \href {https://arxiv.org/abs/2205.00159} {Svtr: Scene text recognition with a single visual model}.
\newblock \emph{Preprint}, arXiv:2205.00159.

\bibitem[{Du et~al.(2020)Du, Li, Guo, Yin, Liu, Zhou, Bai, Yu, Yang, Dang, and Wang}]{DBLP:journals/corr/abs-2009-09941}
Yuning Du, Chenxia Li, Ruoyu Guo, Xiaoting Yin, Weiwei Liu, Jun Zhou, Yifan Bai, Zilin Yu, Yehua Yang, Qingqing Dang, and Haoshuang Wang. 2020.
\newblock \href {https://arxiv.org/abs/2009.09941} {{PP-OCR:} {A} practical ultra lightweight {OCR} system}.
\newblock \emph{CoRR}, abs/2009.09941.

\bibitem[{Guillaume~Jaume(2019)}]{jaume2019}
Jean-Philippe~Thiran Guillaume~Jaume, Hazim Kemal~Ekenel. 2019.
\newblock Funsd: A dataset for form understanding in noisy scanned documents.
\newblock In \emph{Accepted to ICDAR-OST}.

\bibitem[{Huang et~al.(2019)Huang, Chen, He, Bai, Karatzas, Lu, and Jawahar}]{Huang_2019}
Zheng Huang, Kai Chen, Jianhua He, Xiang Bai, Dimosthenis Karatzas, Shijian Lu, and C.~V. Jawahar. 2019.
\newblock \href {https://doi.org/10.1109/icdar.2019.00244} {Icdar2019 competition on scanned receipt ocr and information extraction}.
\newblock In \emph{2019 International Conference on Document Analysis and Recognition (ICDAR)}. IEEE.

\bibitem[{Karatzas et~al.(2015)Karatzas, Gomez-Bigorda, Nicolaou, Ghosh, Bagdanov, Iwamura, Matas, Neumann, Chandrasekhar, Lu, Shafait, Uchida, and Valveny}]{ic15}
Dimosthenis Karatzas, Lluis Gomez-Bigorda, Anguelos Nicolaou, Suman Ghosh, Andrew Bagdanov, Masakazu Iwamura, Jiri Matas, Lukas Neumann, {Vijay Ramaseshan} Chandrasekhar, Shijian Lu, Faisal Shafait, Seiichi Uchida, and Ernest Valveny. 2015.
\newblock \href {https://doi.org/10.1109/ICDAR.2015.7333942} {Icdar 2015 competition on robust reading}.
\newblock In \emph{13th IAPR International Conference on Document Analysis and Recognition, ICDAR 2015 - Conference Proceedings}, Proceedings of the International Conference on Document Analysis and Recognition, ICDAR, pages 1156--1160, United States. IEEE Computer Society.
\newblock Publisher Copyright: {\textcopyright} 2015 IEEE.; 13th International Conference on Document Analysis and Recognition, ICDAR 2015 ; Conference date: 23-08-2015 Through 26-08-2015.

\bibitem[{Karatzas et~al.(2013)Karatzas, Shafait, Uchida, Iwamura, Bigorda, Mestre, Mas, Mota, Almazan, and {De Las Heras}}]{ic13}
Dimosthenis Karatzas, Faisal Shafait, Seiichi Uchida, Masakazu Iwamura, {Lluis Gomez I.} Bigorda, {Sergi Robles} Mestre, Joan Mas, {David Fernandez} Mota, {Jon Almazan} Almazan, and {Lluis Pere} {De Las Heras}. 2013.
\newblock \href {https://doi.org/10.1109/ICDAR.2013.221} {Icdar 2013 robust reading competition}.
\newblock \emph{Proceedings of the International Conference on Document Analysis and Recognition, ICDAR}, pages 1484--1493.
\newblock Copyright: Copyright 2013 Elsevier B.V., All rights reserved.; 12th International Conference on Document Analysis and Recognition, ICDAR 2013 ; Conference date: 25-08-2013 Through 28-08-2013.

\bibitem[{Kim et~al.(2022)Kim, Hong, Yim, Nam, Park, Yim, Hwang, Yun, Han, and Park}]{kim2022donut}
Geewook Kim, Teakgyu Hong, Moonbin Yim, JeongYeon Nam, Jinyoung Park, Jinyeong Yim, Wonseok Hwang, Sangdoo Yun, Dongyoon Han, and Seunghyun Park. 2022.
\newblock Ocr-free document understanding transformer.
\newblock In \emph{European Conference on Computer Vision (ECCV)}.

\bibitem[{Kim et~al.(2019)Kim, Lim, Park, and Cho}]{kim2019korean}
Jin-Hwa Kim, Soohyun Lim, Jaesun Park, and Hansu Cho. 2019.
\newblock \href {https://aiforsocialgood.github.io/neurips2019/accepted/track1/pdfs/44_aisg_neurips2019.pdf} {Korean localization of visual question answering for blind people}.
\newblock In \emph{Proceedings of the AI for Social Good Workshop at NeurIPS}.

\bibitem[{Kim et~al.(2024)Kim, Pertsch, Karamcheti, Xiao, Balakrishna, Nair, Rafailov, Foster, Lam, Sanketi, Vuong, Kollar, Burchfiel, Tedrake, Sadigh, Levine, Liang, and Finn}]{kim24openvla}
{Moo Jin} Kim, Karl Pertsch, Siddharth Karamcheti, Ted Xiao, Ashwin Balakrishna, Suraj Nair, Rafael Rafailov, Ethan Foster, Grace Lam, Pannag Sanketi, Quan Vuong, Thomas Kollar, Benjamin Burchfiel, Russ Tedrake, Dorsa Sadigh, Sergey Levine, Percy Liang, and Chelsea Finn. 2024.
\newblock Openvla: An open-source vision-language-action model.
\newblock \emph{arXiv preprint arXiv:2406.09246}.

\bibitem[{Kwon et~al.(2023)Kwon, Li, Zhuang, Sheng, Zheng, Yu, Gonzalez, Zhang, and Stoica}]{kwon2023efficient}
Woosuk Kwon, Zhuohan Li, Siyuan Zhuang, Ying Sheng, Lianmin Zheng, Cody~Hao Yu, Joseph~E. Gonzalez, Hao Zhang, and Ion Stoica. 2023.
\newblock Efficient memory management for large language model serving with pagedattention.
\newblock In \emph{Proceedings of the ACM SIGOPS 29th Symposium on Operating Systems Principles}.

\bibitem[{Li et~al.(2022{\natexlab{a}})Li, Liu, Guo, Yin, Jiang, Du, Du, Zhu, Lai, Hu, Yu, and Ma}]{li2022ppocrv3attemptsimprovementultra}
Chenxia Li, Weiwei Liu, Ruoyu Guo, Xiaoting Yin, Kaitao Jiang, Yongkun Du, Yuning Du, Lingfeng Zhu, Baohua Lai, Xiaoguang Hu, Dianhai Yu, and Yanjun Ma. 2022{\natexlab{a}}.
\newblock \href {https://arxiv.org/abs/2206.03001} {Pp-ocrv3: More attempts for the improvement of ultra lightweight ocr system}.
\newblock \emph{Preprint}, arXiv:2206.03001.

\bibitem[{Li et~al.(2022{\natexlab{b}})Li, Li, Xiong, and Hoi}]{li2022blip}
Junnan Li, Dongxu Li, Caiming Xiong, and Steven Hoi. 2022{\natexlab{b}}.
\newblock Blip: Bootstrapping language-image pre-training for unified vision-language understanding and generation.
\newblock In \emph{ICML}.

\bibitem[{Li et~al.(2022{\natexlab{c}})Li, Li, Xiong, and Hoi}]{Li2022BLIPBL}
Junnan Li, Dongxu Li, Caiming Xiong, and Steven C.~H. Hoi. 2022{\natexlab{c}}.
\newblock \href {https://api.semanticscholar.org/CorpusID:246411402} {Blip: Bootstrapping language-image pre-training for unified vision-language understanding and generation}.
\newblock In \emph{International Conference on Machine Learning}.

\bibitem[{Li et~al.(2021)Li, Lv, Cui, Lu, Florencio, Zhang, Li, and Wei}]{li2021trocr}
Minghao Li, Tengchao Lv, Lei Cui, Yijuan Lu, Dinei Florencio, Cha Zhang, Zhoujun Li, and Furu Wei. 2021.
\newblock \href {https://arxiv.org/abs/2109.10282} {Trocr: Transformer-based optical character recognition with pre-trained models}.
\newblock \emph{Preprint}, arXiv:2109.10282.

\bibitem[{Liao et~al.(2020)Liao, Wan, Yao, Chen, and Bai}]{liao2020real}
Minghui Liao, Zhaoyi Wan, Cong Yao, Kai Chen, and Xiang Bai. 2020.
\newblock Real-time scene text detection with differentiable binarization.
\newblock In \emph{Proc. AAAI}.

\bibitem[{Liao et~al.(2022)Liao, Zou, Wan, Yao, and Bai}]{liao2022real}
Minghui Liao, Zhisheng Zou, Zhaoyi Wan, Cong Yao, and Xiang Bai. 2022.
\newblock Real-time scene text detection with differentiable binarization and adaptive scale fusion.
\newblock \emph{IEEE Transactions on Pattern Analysis and Machine Intelligence}.

\bibitem[{Liu et~al.(2023)Liu, Li, Wu, and Lee}]{NEURIPS2023_6dcf277e}
Haotian Liu, Chunyuan Li, Qingyang Wu, and Yong~Jae Lee. 2023.
\newblock \href {https://proceedings.neurips.cc/paper_files/paper/2023/file/6dcf277ea32ce3288914faf369fe6de0-Paper-Conference.pdf} {Visual instruction tuning}.
\newblock In \emph{Advances in Neural Information Processing Systems}, volume~36, pages 34892--34916. Curran Associates, Inc.

\bibitem[{Liu et~al.(2019)Liu, Ott, Goyal, Du, Joshi, Chen, Levy, Lewis, Zettlemoyer, and Stoyanov}]{DBLP:journals/corr/abs-1907-11692}
Yinhan Liu, Myle Ott, Naman Goyal, Jingfei Du, Mandar Joshi, Danqi Chen, Omer Levy, Mike Lewis, Luke Zettlemoyer, and Veselin Stoyanov. 2019.
\newblock \href {https://arxiv.org/abs/1907.11692} {Roberta: {A} robustly optimized {BERT} pretraining approach}.
\newblock \emph{CoRR}, abs/1907.11692.

\bibitem[{Liu et~al.(2024)Liu, Li, Huang, Yang, Yu, Li, Yin, Liu, Jin, and Bai}]{Liu_2024}
Yuliang Liu, Zhang Li, Mingxin Huang, Biao Yang, Wenwen Yu, Chunyuan Li, Xu-Cheng Yin, Cheng-Lin Liu, Lianwen Jin, and Xiang Bai. 2024.
\newblock \href {https://doi.org/10.1007/s11432-024-4235-6} {Ocrbench: on the hidden mystery of ocr in large multimodal models}.
\newblock \emph{Science China Information Sciences}, 67(12).

\bibitem[{Loshchilov and Hutter(2019)}]{loshchilov2018decoupled}
Ilya Loshchilov and Frank Hutter. 2019.
\newblock \href {https://openreview.net/forum?id=Bkg6RiCqY7} {Decoupled weight decay regularization}.
\newblock In \emph{International Conference on Learning Representations}.

\bibitem[{Masry et~al.(2022{\natexlab{a}})Masry, Long, Tan, Joty, and Hoque}]{masry-etal-2022-chartqa}
Ahmed Masry, Do~Long, Jia~Qing Tan, Shafiq Joty, and Enamul Hoque. 2022{\natexlab{a}}.
\newblock \href {https://doi.org/10.18653/v1/2022.findings-acl.177} {{C}hart{QA}: A benchmark for question answering about charts with visual and logical reasoning}.
\newblock In \emph{Findings of the Association for Computational Linguistics: ACL 2022}, pages 2263--2279, Dublin, Ireland. Association for Computational Linguistics.

\bibitem[{Masry et~al.(2022{\natexlab{b}})Masry, Long, Tan, Joty, and Hoque}]{masry2022chartqabenchmarkquestionanswering}
Ahmed Masry, Do~Xuan Long, Jia~Qing Tan, Shafiq Joty, and Enamul Hoque. 2022{\natexlab{b}}.
\newblock \href {https://arxiv.org/abs/2203.10244} {Chartqa: A benchmark for question answering about charts with visual and logical reasoning}.
\newblock \emph{Preprint}, arXiv:2203.10244.

\bibitem[{Mathew et~al.(2021)Mathew, Karatzas, and Jawahar}]{9423358}
Minesh Mathew, Dimosthenis Karatzas, and C.~V. Jawahar. 2021.
\newblock \href {https://doi.org/10.1109/WACV48630.2021.00225} {Docvqa: A dataset for vqa on document images}.
\newblock In \emph{2021 IEEE Winter Conference on Applications of Computer Vision (WACV)}, pages 2199--2208.

\bibitem[{Mishra et~al.(2012)Mishra, Alahari, and Jawahar}]{iii5k}
Anand Mishra, Karteek Alahari, and C.~Jawahar. 2012.
\newblock \href {https://doi.org/10.5244/C.26.127} {Scene text recognition using higher order language priors}.

\bibitem[{Muennighoff et~al.(2025)Muennighoff, Yang, Shi, Li, Fei-Fei, Hajishirzi, Zettlemoyer, Liang, Candès, and Hashimoto}]{muennighoff2025s1simpletesttimescaling}
Niklas Muennighoff, Zitong Yang, Weijia Shi, Xiang~Lisa Li, Li~Fei-Fei, Hannaneh Hajishirzi, Luke Zettlemoyer, Percy Liang, Emmanuel Candès, and Tatsunori Hashimoto. 2025.
\newblock \href {https://arxiv.org/abs/2501.19393} {s1: Simple test-time scaling}.
\newblock \emph{Preprint}, arXiv:2501.19393.

\bibitem[{Nacson et~al.(2024)Nacson, Aberdam, Ganz, Avraham, Golts, Kittenplon, Mazor, and Litman}]{nacson2024docvlmmakevlmefficient}
Mor~Shpigel Nacson, Aviad Aberdam, Roy Ganz, Elad~Ben Avraham, Alona Golts, Yair Kittenplon, Shai Mazor, and Ron Litman. 2024.
\newblock \href {https://arxiv.org/abs/2412.08746} {Docvlm: Make your vlm an efficient reader}.
\newblock \emph{Preprint}, arXiv:2412.08746.

\bibitem[{OpenAI et~al.(2024)OpenAI, :, Jaech, Kalai, Lerer, Richardson, El-Kishky, Low, Helyar, Madry, Beutel, Carney, Iftimie, Karpenko, Passos, Neitz, Prokofiev, Wei, Tam, Bennett, Kumar, Saraiva, Vallone, Duberstein, Kondrich, Mishchenko, Applebaum, Jiang, Nair, Zoph, Ghorbani, Rossen, Sokolowsky, Barak, McGrew, Minaiev, Hao, Baker, Houghton, McKinzie, Eastman, Lugaresi, Bassin, Hudson, Li, de~Bourcy, Voss, Shen, Zhang, Koch, Orsinger, Hesse, Fischer, Chan, Roberts, Kappler, Levy, Selsam, Dohan, Farhi, Mely, Robinson, Tsipras, Li, Oprica, Freeman, Zhang, Wong, Proehl, Cheung, Mitchell, Wallace, Ritter, Mays, Wang, Such, Raso, Leoni, Tsimpourlas, Song, von Lohmann, Sulit, Salmon, Parascandolo, Chabot, Zhao, Brockman, Leclerc, Salman, Bao, Sheng, Andrin, Bagherinezhad, Ren, Lightman, Chung, Kivlichan, O'Connell, Osband, Gilaberte, Akkaya, Kostrikov, Sutskever, Kofman, Pachocki, Lennon, Wei, Harb, Twore, Feng, Yu, Weng, Tang, Yu, Candela, Palermo, Parish, Heidecke, Hallman, Rizzo, Gordon, Uesato, Ward,
  Huizinga, Wang, Chen, Xiao, Singhal, Nguyen, Cobbe, Shi, Wood, Rimbach, Gu-Lemberg, Liu, Lu, Stone, Yu, Ahmad, Yang, Liu, Maksin, Ho, Fedus, Weng, Li, McCallum, Held, Kuhn, Kondraciuk, Kaiser, Metz, Boyd, Trebacz, Joglekar, Chen, Tintor, Meyer, Jones, Kaufer, Schwarzer, Shah, Yatbaz, Guan, Xu, Yan, Glaese, Chen, Lampe, Malek, Wang, Fradin, McClay, Pavlov, Wang, Wang, Murati, Bavarian, Rohaninejad, McAleese, Chowdhury, Chowdhury, Ryder, Tezak, Brown, Nachum, Boiko, Murk, Watkins, Chao, Ashbourne, Izmailov, Zhokhov, Dias, Arora, Lin, Lopes, Gaon, Miyara, Leike, Hwang, Garg, Brown, James, Shu, Cheu, Greene, Jain, Altman, Toizer, Toyer, Miserendino, Agarwal, Hernandez, Baker, McKinney, Yan, Zhao, Hu, Santurkar, Chaudhuri, Zhang, Fu, Papay, Lin, Balaji, Sanjeev, Sidor, Broda, Clark, Wang, Gordon, Sanders, Patwardhan, Sottiaux, Degry, Dimson, Zheng, Garipov, Stasi, Bansal, Creech, Peterson, Eloundou, Qi, Kosaraju, Monaco, Pong, Fomenko, Zheng, Zhou, McCabe, Zaremba, Dubois, Lu, Chen, Cha, Bai, He, Zhang, Wang,
  Shao, and Li}]{openai2024openaio1card}
OpenAI, :, Aaron Jaech, Adam Kalai, Adam Lerer, Adam Richardson, Ahmed El-Kishky, Aiden Low, Alec Helyar, Aleksander Madry, Alex Beutel, Alex Carney, Alex Iftimie, Alex Karpenko, Alex~Tachard Passos, Alexander Neitz, Alexander Prokofiev, Alexander Wei, Allison Tam, and 244 others. 2024.
\newblock \href {https://arxiv.org/abs/2412.16720} {Openai o1 system card}.
\newblock \emph{Preprint}, arXiv:2412.16720.

\bibitem[{Paszke et~al.(2019)Paszke, Gross, Massa, Lerer, Bradbury, Chanan, Killeen, Lin, Gimelshein, Antiga, Desmaison, Köpf, Yang, DeVito, Raison, Tejani, Chilamkurthy, Steiner, Fang, Bai, and Chintala}]{paszke2019pytorchimperativestylehighperformance}
Adam Paszke, Sam Gross, Francisco Massa, Adam Lerer, James Bradbury, Gregory Chanan, Trevor Killeen, Zeming Lin, Natalia Gimelshein, Luca Antiga, Alban Desmaison, Andreas Köpf, Edward Yang, Zach DeVito, Martin Raison, Alykhan Tejani, Sasank Chilamkurthy, Benoit Steiner, Lu~Fang, and 2 others. 2019.
\newblock \href {https://arxiv.org/abs/1912.01703} {Pytorch: An imperative style, high-performance deep learning library}.
\newblock \emph{Preprint}, arXiv:1912.01703.

\bibitem[{Peng et~al.(2023)Peng, Lee, Wang, Balasubramaniyan, and Chau}]{peng2023high}
Anthony Peng, Seongmin Lee, Xiaojing Wang, Rajarajeswari~Raji Balasubramaniyan, and Duen~Horng Chau. 2023.
\newblock High-performance transformers for table structure recognition need early convolutions.
\newblock In \emph{NeurIPS 2023 Second Table Representation Learning Workshop}.

\bibitem[{Peng et~al.(2024{\natexlab{a}})Peng, Lee, Wang, Balasubramaniyan, and Chau}]{peng2024self}
ShengYun Peng, Seongmin Lee, Xiaojing Wang, Rajarajeswari Balasubramaniyan, and Duen~Horng Chau. 2024{\natexlab{a}}.
\newblock Self-supervised pretraining for table structure recognition transformer.
\newblock \emph{arXiv preprint}.

\bibitem[{Peng et~al.(2024{\natexlab{b}})Peng, Lee, Wang, Balasubramaniyan, and Chau}]{peng2024unitable}
ShengYun Peng, Seongmin Lee, Xiaojing Wang, Rajarajeswari Balasubramaniyan, and Duen~Horng Chau. 2024{\natexlab{b}}.
\newblock Unitable: Towards a unified framework for table structure recognition via self-supervised pretraining.
\newblock \emph{arXiv preprint}.

\bibitem[{Pfitzmann et~al.(2022)Pfitzmann, Auer, Dolfi, Nassar, and Staar}]{doclaynet2022}
Birgit Pfitzmann, Christoph Auer, Michele Dolfi, Ahmed~S Nassar, and Peter W~J Staar. 2022.
\newblock \href {https://doi.org/10.1145/3534678.353904} {Doclaynet: A large human-annotated dataset for document-layout analysis}.

\bibitem[{Phan et~al.(2013)Phan, Shivakumara, Tian, and Tan}]{svtp}
Trung Phan, Palaiahnakote Shivakumara, Shuangxuan Tian, and Chew~Lim Tan. 2013.
\newblock \href {https://doi.org/10.1109/ICCV.2013.76} {Recognizing text with perspective distortion in natural scenes}.
\newblock pages 569--576.

\bibitem[{{Pixparse}(2024)}]{pixparse_idl_wds}
{Pixparse}. 2024.
\newblock {idl-wds} dataset.
\newblock \url{https://huggingface.co/datasets/pixparse/idl-wds}.
\newblock Accessed: 2025-04-01.

\bibitem[{Rang et~al.(2024{\natexlab{a}})Rang, Bi, Liu, Wang, and Han}]{rang2024empiricalstudyscalinglaw}
Miao Rang, Zhenni Bi, Chuanjian Liu, Yunhe Wang, and Kai Han. 2024{\natexlab{a}}.
\newblock \href {https://arxiv.org/abs/2401.00028} {An empirical study of scaling law for ocr}.
\newblock \emph{Preprint}, arXiv:2401.00028.

\bibitem[{Rang et~al.(2024{\natexlab{b}})Rang, Bi, Liu, Wang, and Han}]{Rang_2024_CVPR}
Miao Rang, Zhenni Bi, Chuanjian Liu, Yunhe Wang, and Kai Han. 2024{\natexlab{b}}.
\newblock An empirical study of scaling law for scene text recognition.
\newblock In \emph{Proceedings of the IEEE/CVF Conference on Computer Vision and Pattern Recognition (CVPR)}, pages 15619--15629.

\bibitem[{Risnumawan et~al.(2014)Risnumawan, Shivakumara, Chan, and Tan}]{cute}
Anhar Risnumawan, Palaiahnakote Shivakumara, Chee~Seng Chan, and Chew~Lim Tan. 2014.
\newblock \href {https://api.semanticscholar.org/CorpusID:15559857} {A robust arbitrary text detection system for natural scene images}.
\newblock \emph{Expert Syst. Appl.}, 41:8027--8048.

\bibitem[{Shi et~al.(2017)Shi, Bai, and Yao}]{7801919}
Baoguang Shi, Xiang Bai, and Cong Yao. 2017.
\newblock \href {https://doi.org/10.1109/TPAMI.2016.2646371} {An end-to-end trainable neural network for image-based sequence recognition and its application to scene text recognition}.
\newblock \emph{IEEE Transactions on Pattern Analysis and Machine Intelligence}, 39(11):2298--2304.

\bibitem[{Singh et~al.(2019)Singh, Natarjan, Shah, Jiang, Chen, Parikh, and Rohrbach}]{singh2019towards}
Amanpreet Singh, Vivek Natarjan, Meet Shah, Yu~Jiang, Xinlei Chen, Devi Parikh, and Marcus Rohrbach. 2019.
\newblock Towards vqa models that can read.
\newblock In \emph{Proceedings of the IEEE Conference on Computer Vision and Pattern Recognition}, pages 8317--8326.

\bibitem[{Smock et~al.(2022)Smock, Pesala, and Abraham}]{smock2022pubtables}
Brandon Smock, Rohith Pesala, and Robin Abraham. 2022.
\newblock Pub{T}ables-1{M}: Towards comprehensive table extraction from unstructured documents.
\newblock In \emph{Proceedings of the IEEE/CVF Conference on Computer Vision and Pattern Recognition (CVPR)}, pages 4634--4642.

\bibitem[{Tang et~al.(2024)Tang, Liu, Ye, Lu, Wei, Lin, Li, Mahmood, Feng, Zhao, Wang, Liu, Liu, Bai, and Huang}]{tang2024mtvqa}
Jingqun Tang, Qi~Liu, Yongjie Ye, Jinghui Lu, Shu Wei, Chunhui Lin, Wanqing Li, Mohamad Fitri Faiz~Bin Mahmood, Hao Feng, Zhen Zhao, Yanjie Wang, Yuliang Liu, Hao Liu, Xiang Bai, and Can Huang. 2024.
\newblock \href {https://arxiv.org/abs/2405.11985} {Mtvqa: Benchmarking multilingual text-centric visual question answering}.
\newblock \emph{Preprint}, arXiv:2405.11985.

\bibitem[{Team et~al.(2024)Team, Anil, Borgeaud, Alayrac, Yu, Soricut, Schalkwyk, Dai, Hauth, Millican, Silver, Johnson, Antonoglou, Schrittwieser, Glaese, Chen, Pitler, Lillicrap, Lazaridou, Firat, Molloy, Isard, Barham, Hennigan, Lee, Viola, Reynolds, Xu, Doherty, Collins, Meyer, Rutherford, Moreira, Ayoub, Goel, Krawczyk, Du, Chi, Cheng, Ni, Shah, Kane, Chan, Faruqui, Severyn, Lin, Li, Cheng, Ittycheriah, Mahdieh, Chen, Sun, Tran, Bagri, Lakshminarayanan, Liu, Orban, Güra, Zhou, Song, Boffy, Ganapathy, Zheng, Choe, Ágoston Weisz, Zhu, Lu, Gopal, Kahn, Kula, Pitman, Shah, Taropa, Merey, Baeuml, Chen, Shafey, Zhang, Sercinoglu, Tucker, Piqueras, Krikun, Barr, Savinov, Danihelka, Roelofs, White, Andreassen, von Glehn, Yagati, Kazemi, Gonzalez, Khalman, Sygnowski, Frechette, Smith, Culp, Proleev, Luan, Chen, Lottes, Schucher, Lebron, Rrustemi, Clay, Crone, Kocisky, Zhao, Perz, Yu, Howard, Bloniarz, Rae, Lu, Sifre, Maggioni, Alcober, Garrette, Barnes, Thakoor, Austin, Barth-Maron, Wong, Joshi, Chaabouni,
  Fatiha, Ahuja, Tomar, Senter, Chadwick, Kornakov, Attaluri, Iturrate, Liu, Li, Cogan, Chen, Jia, Gu, Zhang, Grimstad, Hartman, Garcia, Pillai, Devlin, Laskin, de~Las~Casas, Valter, Tao, Blanco, Badia, Reitter, Chen, Brennan, Rivera, Brin, Iqbal, Surita, Labanowski, Rao, Winkler, Parisotto, Gu, Olszewska, Addanki, Miech, Louis, Teplyashin, Brown, Catt, Balaguer, Xiang, Wang, Ashwood, Briukhov, Webson, Ganapathy, Sanghavi, Kannan, Chang, Stjerngren, Djolonga, Sun, Bapna, Aitchison, Pejman, Michalewski, Yu, Wang, Love, Ahn, Bloxwich, Han, Humphreys, Sellam, Bradbury, Godbole, Samangooei, Damoc, Kaskasoli, Arnold, Vasudevan, Agrawal, Riesa, Lepikhin, Tanburn, Srinivasan, Lim, Hodkinson, Shyam, Ferret, Hand, Garg, Paine, Li, Li, Giang, Neitz, Abbas, York, Reid, Cole, Chowdhery, Das, Rogozińska, Nikolaev, Sprechmann, Nado, Zilka, Prost, He, Monteiro, Mishra, Welty, Newlan, Jia, Allamanis, Hu, de~Liedekerke, Gilmer, Saroufim, Rijhwani, Hou, Shrivastava, Baddepudi, Goldin, Ozturel, Cassirer, Xu, Sohn, Sachan,
  Amplayo, Swanson, Petrova, Narayan, Guez, Brahma, Landon, Patel, Zhao, Villela, Wang, Jia, Rahtz, Giménez, Yeung, Keeling, Georgiev, Mincu, Wu, Haykal, Saputro, Vodrahalli, Qin, Cankara, Sharma, Fernando, Hawkins, Neyshabur, Kim, Hutter, Agrawal, Castro-Ros, van~den Driessche, Wang, Yang, yiin Chang, Komarek, McIlroy, Lučić, Zhang, Farhan, Sharman, Natsev, Michel, Bansal, Qiao, Cao, Shakeri, Butterfield, Chung, Rubenstein, Agrawal, Mensch, Soparkar, Lenc, Chung, Pope, Maggiore, Kay, Jhakra, Wang, Maynez, Phuong, Tobin, Tacchetti, Trebacz, Robinson, Katariya, Riedel, Bailey, Xiao, Ghelani, Aroyo, Slone, Houlsby, Xiong, Yang, Gribovskaya, Adler, Wirth, Lee, Li, Kagohara, Pavagadhi, Bridgers, Bortsova, Ghemawat, Ahmed, Liu, Powell, Bolina, Iinuma, Zablotskaia, Besley, Chung, Dozat, Comanescu, Si, Greer, Su, Polacek, Kaufman, Tokumine, Hu, Buchatskaya, Miao, Elhawaty, Siddhant, Tomasev, Xing, Greer, Miller, Ashraf, Roy, Zhang, Ma, Filos, Besta, Blevins, Klimenko, Yeh, Changpinyo, Mu, Chang, Pajarskas, Muir,
  Cohen, Lan, Haridasan, Marathe, Hansen, Douglas, Samuel, Wang, Austin, Lan, Jiang, Chiu, Lorenzo, Sjösund, Cevey, Gleicher, Avrahami, Boral, Srinivasan, Selo, May, Aisopos, Hussenot, Soares, Baumli, Chang, Recasens, Caine, Pritzel, Pavetic, Pardo, Gergely, Frye, Ramasesh, Horgan, Badola, Kassner, Roy, Dyer, Campos, Tomala, Tang, Badawy, White, Mustafa, Lang, Jindal, Vikram, Gong, Caelles, Hemsley, Thornton, Feng, Stokowiec, Zheng, Thacker, Çağlar Ünlü, Zhang, Saleh, Svensson, Bileschi, Patil, Anand, Ring, Tsihlas, Vezer, Selvi, Shevlane, Rodriguez, Kwiatkowski, Daruki, Rong, Dafoe, FitzGerald, Gu-Lemberg, Khan, Hendricks, Pellat, Feinberg, Cobon-Kerr, Sainath, Rauh, Hashemi, Ives, Hasson, Noland, Cao, Byrd, Hou, Wang, Sottiaux, Paganini, Lespiau, Moufarek, Hassan, Shivakumar, van Amersfoort, Mandhane, Joshi, Goyal, Tung, Brock, Sheahan, Misra, Li, Rakićević, Dehghani, Liu, Mittal, Oh, Noury, Sezener, Huot, Lamm, Cao, Chen, Mudgal, Stella, Brooks, Vasudevan, Liu, Chain, Melinkeri, Cohen, Wang,
  Seymore, Zubkov, Goel, Yue, Krishnakumaran, Albert, Hurley, Sano, Mohananey, Joughin, Filonov, Kępa, Eldawy, Lim, Rishi, Badiezadegan, Bos, Chang, Jain, Padmanabhan, Puttagunta, Krishna, Baker, Kalb, Bedapudi, Kurzrok, Lei, Yu, Litvin, Zhou, Wu, Sobell, Siciliano, Papir, Neale, Bragagnolo, Toor, Chen, Anklin, Wang, Feng, Gholami, Ling, Liu, Walter, Moghaddam, Kishore, Adamek, Mercado, Mallinson, Wandekar, Cagle, Ofek, Garrido, Lombriser, Mukha, Sun, Mohammad, Matak, Qian, Peswani, Janus, Yuan, Schelin, David, Garg, He, Duzhyi, Älgmyr, Lottaz, Li, Yadav, Xu, Chinien, Shivanna, Chuklin, Li, Spadine, Wolfe, Mohamed, Das, Dai, He, von Dincklage, Upadhyay, Maurya, Chi, Krause, Salama, Rabinovitch, M, Selvan, Dektiarev, Ghiasi, Guven, Gupta, Liu, Sharma, Shtacher, Paul, Akerlund, Aubet, Huang, Zhu, Zhu, Teixeira, Fritze, Bertolini, Marinescu, Bölle, Paulus, Gupta, Latkar, Chang, Sanders, Wilson, Wu, Tan, Thiet, Doshi, Lall, Mishra, Chen, Luong, Benjamin, Lee, Andrejczuk, Rabiej, Ranjan, Styrc, Yin, Simon,
  Harriott, Bansal, Robsky, Bacon, Greene, Mirylenka, Zhou, Sarvana, Goyal, Andermatt, Siegler, Horn, Israel, Pongetti, Chen, Selvatici, Silva, Wang, Tolins, Guu, Yogev, Cai, Agostini, Shah, Nguyen, Donnaile, Pereira, Friso, Stambler, Kurzrok, Kuang, Romanikhin, Geller, Yan, Jang, Lee, Fica, Malmi, Tan, Banica, Balle, Pham, Huang, Avram, Shi, Singh, Hidey, Ahuja, Saxena, Dooley, Potharaju, O'Neill, Gokulchandran, Foley, Zhao, Dusenberry, Liu, Mehta, Kotikalapudi, Safranek-Shrader, Goodman, Kessinger, Globen, Kolhar, Gorgolewski, Ibrahim, Song, Eichenbaum, Brovelli, Potluri, Lahoti, Baetu, Ghorbani, Chen, Crawford, Pal, Sridhar, Gurita, Mujika, Petrovski, Cedoz, Li, Chen, Santo, Goyal, Punjabi, Kappaganthu, Kwak, LV, Velury, Choudhury, Hall, Shah, Figueira, Thomas, Lu, Zhou, Kumar, Jurdi, Chikkerur, Ma, Yu, Kwak, Ähdel, Rajayogam, Choma, Liu, Barua, Ji, Park, Hellendoorn, Bailey, Bilal, Zhou, Khatir, Sutton, Rzadkowski, Macintosh, Shagin, Medina, Liang, Zhou, Shah, Bi, Dankovics, Banga, Lehmann, Bredesen,
  Lin, Hoffmann, Lai, Chung, Yang, Balani, Bražinskas, Sozanschi, Hayes, Alcalde, Makarov, Chen, Stella, Snijders, Mandl, Kärrman, Nowak, Wu, Dyck, Vaidyanathan, R, Mallet, Rudominer, Johnston, Mittal, Udathu, Christensen, Verma, Irving, Santucci, Elsayed, Davoodi, Georgiev, Tenney, Hua, Cideron, Leurent, Alnahlawi, Georgescu, Wei, Zheng, Scandinaro, Jiang, Snoek, Sundararajan, Wang, Ontiveros, Karo, Cole, Rajashekhar, Tumeh, Ben-David, Jain, Uesato, Datta, Bunyan, Wu, Zhang, Stanczyk, Zhang, Steiner, Naskar, Azzam, Johnson, Paszke, Chiu, Elias, Mohiuddin, Muhammad, Miao, Lee, Vieillard, Park, Zhang, Stanway, Garmon, Karmarkar, Dong, Lee, Kumar, Zhou, Evens, Isaac, Irving, Loper, Fink, Arkatkar, Chen, Shafran, Petrychenko, Chen, Jia, Levskaya, Zhu, Grabowski, Mao, Magni, Yao, Snaider, Casagrande, Palmer, Suganthan, Castaño, Giannoumis, Kim, Rybiński, Sreevatsa, Prendki, Soergel, Goedeckemeyer, Gierke, Jafari, Gaba, Wiesner, Wright, Wei, Vashisht, Kulizhskaya, Hoover, Le, Li, Iwuanyanwu, Liu, Ramirez,
  Khorlin, Cui, LIN, Wu, Aguilar, Pallo, Chakladar, Perng, Abellan, Zhang, Dasgupta, Kushman, Penchev, Repina, Wu, van~der Weide, Ponnapalli, Kaplan, Simsa, Li, Dousse, Yang, Piper, Ie, Pasumarthi, Lintz, Vijayakumar, Andor, Valenzuela, Lui, Paduraru, Peng, Lee, Zhang, Greene, Nguyen, Kurylowicz, Hardin, Dixon, Janzer, Choo, Feng, Zhang, Singhal, Du, McKinnon, Antropova, Bolukbasi, Keller, Reid, Finchelstein, Raad, Crocker, Hawkins, Dadashi, Gaffney, Franko, Bulanova, Leblond, Chung, Askham, Cobo, Xu, Fischer, Xu, Sorokin, Alberti, Lin, Evans, Dimitriev, Forbes, Banarse, Tung, Omernick, Bishop, Sterneck, Jain, Xia, Amid, Piccinno, Wang, Banzal, Mankowitz, Polozov, Krakovna, Brown, Bateni, Duan, Firoiu, Thotakuri, Natan, Geist, tan Girgin, Li, Ye, Roval, Tojo, Kwong, Lee-Thorp, Yew, Sinopalnikov, Ramos, Mellor, Sharma, Wu, Miller, Sonnerat, Vnukov, Greig, Beattie, Caveness, Bai, Eisenschlos, Korchemniy, Tsai, Jasarevic, Kong, Dao, Zheng, Liu, Yang, Zhu, Teh, Sanmiya, Gladchenko, Trdin, Toyama, Rosen, Tavakkol,
  Xue, Elkind, Woodman, Carpenter, Papamakarios, Kemp, Kafle, Grunina, Sinha, Talbert, Wu, Owusu-Afriyie, Du, Thornton, Pont-Tuset, Narayana, Li, Fatehi, Wieting, Ajmeri, Uria, Ko, Knight, Héliou, Niu, Gu, Pang, Li, Levine, Stolovich, Santamaria-Fernandez, Goenka, Yustalim, Strudel, Elqursh, Deck, Lee, Li, Levin, Hoffmann, Holtmann-Rice, Bachem, Arora, Koh, Yeganeh, Põder, Tariq, Sun, Ionita, Seyedhosseini, Tafti, Liu, Gulati, Liu, Ye, Chrzaszcz, Wang, Sethi, Li, Brown, Singh, Fan, Parisi, Stanton, Koverkathu, Choquette-Choo, Li, Lu, Ittycheriah, Shroff, Varadarajan, Bahargam, Willoughby, Gaddy, Desjardins, Cornero, Robenek, Mittal, Albrecht, Shenoy, Moiseev, Jacobsson, Ghaffarkhah, Rivière, Walton, Crepy, Parrish, Zhou, Farabet, Radebaugh, Srinivasan, van~der Salm, Fidjeland, Scellato, Latorre-Chimoto, Klimczak-Plucińska, Bridson, de~Cesare, Hudson, Mendolicchio, Walker, Morris, Mauger, Guseynov, Reid, Odoom, Loher, Cotruta, Yenugula, Grewe, Petrushkina, Duerig, Sanchez, Yadlowsky, Shen, Globerson, Webb,
  Dua, Li, Bhupatiraju, Hurt, Qureshi, Agarwal, Shani, Eyal, Khare, Belle, Wang, Tekur, Kale, Wei, Sang, Saeta, Liechty, Sun, Zhao, Lee, Nayak, Fritz, Vuyyuru, Aslanides, Vyas, Wicke, Ma, Eltyshev, Martin, Cate, Manyika, Amiri, Kim, Xiong, Kang, Luisier, Tripuraneni, Madras, Guo, Waters, Wang, Ainslie, Baldridge, Zhang, Pruthi, Bauer, Yang, Mansour, Gelman, Xu, Polovets, Liu, Cai, Chen, Sheng, Xue, Ozair, Angermueller, Li, Sinha, Wang, Wiesinger, Koukoumidis, Tian, Iyer, Gurumurthy, Goldenson, Shah, Blake, Yu, Urbanowicz, Palomaki, Fernando, Durden, Mehta, Momchev, Rahimtoroghi, Georgaki, Raul, Ruder, Redshaw, Lee, Zhou, Jalan, Li, Hechtman, Schuh, Nasr, Milan, Mikulik, Franco, Green, Nguyen, Kelley, Mahendru, Hu, Howland, Vargas, Hui, Bansal, Rao, Ghiya, Wang, Ye, Sarr, Preston, Elish, Li, Kaku, Gupta, Pasupat, Juan, Someswar, M., Chen, Amini, Fabrikant, Chu, Dong, Muthal, Buthpitiya, Jauhari, Hua, Khandelwal, Hitron, Ren, Rinaldi, Drath, Dabush, Jiang, Godhia, Sachs, Chen, Fan, Taitelbaum, Noga, Dai, Wang,
  Liang, Hamer, Ferng, Elkind, Atias, Lee, Listík, Carlen, van~de Kerkhof, Pikus, Zaher, Müller, Zykova, Stefanec, Gatsko, Hirnschall, Sethi, Xu, Ahuja, Tsai, Stefanoiu, Feng, Dhandhania, Katyal, Gupta, Parulekar, Pitta, Zhao, Bhatia, Bhavnani, Alhadlaq, Li, Danenberg, Tu, Pine, Filippova, Ghosh, Limonchik, Urala, Lanka, Clive, Sun, Li, Wu, Hongtongsak, Li, Thakkar, Omarov, Majmundar, Alverson, Kucharski, Patel, Jain, Zabelin, Pelagatti, Kohli, Kumar, Kim, Sankar, Shah, Ramachandruni, Zeng, Bariach, Weidinger, Vu, Andreev, He, Hui, Kashem, Subramanya, Hsiao, Hassabis, Kavukcuoglu, Sadovsky, Le, Strohman, Wu, Petrov, Dean, and Vinyals}]{geminiteam2024geminifamilyhighlycapable}
Gemini Team, Rohan Anil, Sebastian Borgeaud, Jean-Baptiste Alayrac, Jiahui Yu, Radu Soricut, Johan Schalkwyk, Andrew~M. Dai, Anja Hauth, Katie Millican, David Silver, Melvin Johnson, Ioannis Antonoglou, Julian Schrittwieser, Amelia Glaese, Jilin Chen, Emily Pitler, Timothy Lillicrap, Angeliki Lazaridou, and 1331 others. 2024.
\newblock \href {https://arxiv.org/abs/2312.11805} {Gemini: A family of highly capable multimodal models}.
\newblock \emph{Preprint}, arXiv:2312.11805.

\bibitem[{{team-lucid}(2023)}]{team-lucid_trocr_small_korean}
{team-lucid}. 2023.
\newblock trocr-small-korean model.
\newblock \url{https://huggingface.co/team-lucid/trocr-small-korean}.
\newblock Accessed: 2025-04-01.

\bibitem[{Thomas et~al.(2024)Thomas, Gaizauskas, and Lu}]{thomas-etal-2024-leveraging}
Alan Thomas, Robert Gaizauskas, and Haiping Lu. 2024.
\newblock \href {https://aclanthology.org/2024.lt4hala-1.14/} {Leveraging {LLM}s for post-{OCR} correction of historical newspapers}.
\newblock In \emph{Proceedings of the Third Workshop on Language Technologies for Historical and Ancient Languages (LT4HALA) @ LREC-COLING-2024}, pages 116--121, Torino, Italia. ELRA and ICCL.

\bibitem[{Touvron et~al.(2021)Touvron, Cord, Douze, Massa, Sablayrolles, and Jegou}]{pmlr-v139-touvron21a}
Hugo Touvron, Matthieu Cord, Matthijs Douze, Francisco Massa, Alexandre Sablayrolles, and Herve Jegou. 2021.
\newblock \href {https://proceedings.mlr.press/v139/touvron21a.html} {Training data-efficient image transformers \& distillation through attention}.
\newblock In \emph{Proceedings of the 38th International Conference on Machine Learning}, volume 139 of \emph{Proceedings of Machine Learning Research}, pages 10347--10357. PMLR.

\bibitem[{Wang et~al.(2024)Wang, Raman, Sibue, Ma, Babkin, Kaur, Pei, Nourbakhsh, and Liu}]{wang-etal-2024-docllm}
Dongsheng Wang, Natraj Raman, Mathieu Sibue, Zhiqiang Ma, Petr Babkin, Simerjot Kaur, Yulong Pei, Armineh Nourbakhsh, and Xiaomo Liu. 2024.
\newblock \href {https://doi.org/10.18653/v1/2024.acl-long.463} {{D}oc{LLM}: A layout-aware generative language model for multimodal document understanding}.
\newblock In \emph{Proceedings of the 62nd Annual Meeting of the Association for Computational Linguistics (Volume 1: Long Papers)}, pages 8529--8548, Bangkok, Thailand. Association for Computational Linguistics.

\bibitem[{Wang et~al.(2021{\natexlab{a}})Wang, Liu, Jin, Tang, Zhang, Zhang, Wang, Wu, and Cai}]{vies2021}
Jiapeng Wang, Chongyu Liu, Lianwen Jin, Guozhi Tang, Jiaxin Zhang, Shuaitao Zhang, Qianying Wang, Yaqiang Wu, and Mingxiang Cai. 2021{\natexlab{a}}.
\newblock \href {https://doi.org/10.1609/aaai.v35i4.16378} {Towards robust visual information extraction in real world: New dataset and novel solution}.
\newblock \emph{Proceedings of the AAAI Conference on Artificial Intelligence}, 35:2738--2745.

\bibitem[{Wang et~al.(2011)Wang, Babenko, and Belongie}]{10.1109/ICCV.2011.6126402}
Kai Wang, Boris Babenko, and Serge Belongie. 2011.
\newblock \href {https://doi.org/10.1109/ICCV.2011.6126402} {End-to-end scene text recognition}.
\newblock In \emph{Proceedings of the 2011 International Conference on Computer Vision}, ICCV '11, page 1457–1464, USA. IEEE Computer Society.

\bibitem[{Wang et~al.(2021{\natexlab{b}})Wang, Xu, Cui, Shang, and Wei}]{wang2021layoutreaderpretrainingtextlayout}
Zilong Wang, Yiheng Xu, Lei Cui, Jingbo Shang, and Furu Wei. 2021{\natexlab{b}}.
\newblock \href {https://arxiv.org/abs/2108.11591} {Layoutreader: Pre-training of text and layout for reading order detection}.
\newblock \emph{Preprint}, arXiv:2108.11591.

\bibitem[{Wei et~al.(2024)Wei, Liu, Chen, Wang, Kong, Xu, Ge, Zhao, Sun, Peng, Han, and Zhang}]{wei2024general}
Haoran Wei, Chenglong Liu, Jinyue Chen, Jia Wang, Lingyu Kong, Yanming Xu, Zheng Ge, Liang Zhao, Jianjian Sun, Yuang Peng, Chunrui Han, and Xiangyu Zhang. 2024.
\newblock \href {https://openreview.net/forum?id=3LOcwfB4JX} {General {OCR} theory: Towards {OCR}-2.0 via a unified end-to-end model}.

\bibitem[{Yang et~al.(2017)Yang, Yumer, Asente, Kraley, Kifer, and Giles}]{8099945}
Xiao Yang, Ersin Yumer, Paul Asente, Mike Kraley, Daniel Kifer, and C.~Lee Giles. 2017.
\newblock \href {https://doi.org/10.1109/CVPR.2017.462} {Learning to extract semantic structure from documents using multimodal fully convolutional neural networks}.
\newblock In \emph{2017 IEEE Conference on Computer Vision and Pattern Recognition (CVPR)}, pages 4342--4351.

\bibitem[{Yang et~al.(2023)Yang, Long, Wang, Song, Zhong, Cheng, Bai, and Yao}]{10203796}
Zhibo Yang, Rujiao Long, Pengfei Wang, Sibo Song, Humen Zhong, Wenqing Cheng, Xiang Bai, and Cong Yao. 2023.
\newblock \href {https://doi.org/10.1109/CVPR52729.2023.01474} {Modeling entities as semantic points for visual information extraction in the wild}.
\newblock In \emph{2023 IEEE/CVF Conference on Computer Vision and Pattern Recognition (CVPR)}, pages 15358--15367.

\bibitem[{Ye et~al.(2022)Ye, Zhang, Zhao, Liu, Du, and Tao}]{ye2022dptextdetrbetterscenetext}
Maoyuan Ye, Jing Zhang, Shanshan Zhao, Juhua Liu, Bo~Du, and Dacheng Tao. 2022.
\newblock \href {https://arxiv.org/abs/2207.04491} {Dptext-detr: Towards better scene text detection with dynamic points in transformer}.
\newblock \emph{Preprint}, arXiv:2207.04491.

\bibitem[{Ye et~al.(2023)Ye, Zhang, Zhao, Liu, Du, and Tao}]{ye2022dptext}
Maoyuan Ye, Jing Zhang, Shanshan Zhao, Juhua Liu, Bo~Du, and Dacheng Tao. 2023.
\newblock Dptext-detr: Towards better scene text detection with dynamic points in transformer.
\newblock In \emph{Proceedings of the AAAI Conference on Artificial Intelligence}, volume~37, pages 3241--3249.

\bibitem[{Yim et~al.(2021)Yim, Kim, Cho, and Park}]{yim2021synthtiger}
Moonbin Yim, Yoonsik Kim, Han-Cheol Cho, and Sungrae Park. 2021.
\newblock Synthtiger: Synthetic text image generator towards better text recognition models.
\newblock In \emph{International Conference on Document Analysis and Recognition}, pages 109--124. Springer.

\bibitem[{Zhao et~al.(2024)Zhao, Quan, Zhu, and Yang}]{zhao2024clip4str}
Shuai Zhao, Ruijie Quan, Linchao Zhu, and Yi~Yang. 2024.
\newblock \href {https://doi.org/10.1109/TIP.2024.3512354} {Clip4str: A simple baseline for scene text recognition with pre-trained vision-language model}.
\newblock \emph{IEEE Transactions on Image Processing}, pages 1--1.

\bibitem[{Zhong et~al.(2019)Zhong, Tang, and Yepes}]{zhong2019publaynetlargestdatasetdocument}
Xu~Zhong, Jianbin Tang, and Antonio~Jimeno Yepes. 2019.
\newblock \href {https://arxiv.org/abs/1908.07836} {Publaynet: largest dataset ever for document layout analysis}.
\newblock \emph{Preprint}, arXiv:1908.07836.

\bibitem[{Zhong et~al.(2020)Zhong, Zheng, Kang, Li, and Yang}]{Zhong_Zheng_Kang_Li_Yang_2020}
Zhun Zhong, Liang Zheng, Guoliang Kang, Shaozi Li, and Yi~Yang. 2020.
\newblock \href {https://doi.org/10.1609/aaai.v34i07.7000} {Random erasing data augmentation}.
\newblock \emph{Proceedings of the AAAI Conference on Artificial Intelligence}, 34(07):13001--13008.

\end{thebibliography}

\appendix

\section{KLOCR Data Details}
\label{sec:app-data}

\subsection{Mixture}

We report the exact datasets used from AIHub in Table~\ref{tab:data_sources}.

\begin{table}[htbp]
  \centering
  \begin{tabular}{c|c}
    \hline
    \textbf{Dataset} & \textbf{Source} \\
    \hline
    Public Administrative Documents & \href{https://www.aihub.or.kr/aihubdata/data/view.do?currMenu=115&topMenu=100&aihubDataSe=data&dataSetSn=88}{Link}\\
    OCR Data (Public Services) & \href{https://www.aihub.or.kr/aihubdata/data/view.do?currMenu=115&topMenu=100&aihubDataSe=data&dataSetSn=71299}{Link}\\
    Finance Documents Data & \href{https://www.aihub.or.kr/aihubdata/data/view.do?currMenu=115&topMenu=100&aihubDataSe=data&dataSetSn=632}{Link}\\
    Korean Font Images & \href{https://aihub.or.kr/aihubdata/data/view.do?currMenu=115&topMenu=100&aihubDataSe=data&dataSetSn=81}{Link}\\
    OCR Data (Handwriting OCR Data) & \href{https://www.aihub.or.kr/aihubdata/data/view.do?currMenu=115&topMenu=100&aihubDataSe=data&dataSetSn=605}{Link}\\
    Various Korean Characters OCR & \href{https://www.aihub.or.kr/aihubdata/data/view.do?currMenu=115&topMenu=100&aihubDataSe=data&dataSetSn=91}{Link}\\
    OCR Data (Financial and Logistics) & \href{https://www.aihub.or.kr/aihubdata/data/view.do?currMenu=115&topMenu=100&aihubDataSe=data&dataSetSn=71301}{Link}\\
    \hline
  \end{tabular}
  \caption{\label{tab:data_sources} AI Hub data sources in the KLOCR data mixture.}
\end{table}

\subsection{Data Processing}

\begin{figure}[t]
    \centering
    \includegraphics[width=0.48\textwidth]{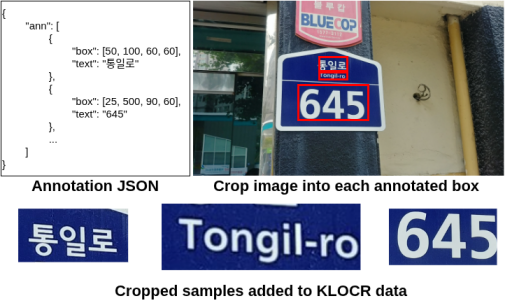}
    \caption{KLOCR data processing.}
    \label{fig:klocr-processing}
\end{figure}

Figure~\ref{fig:klocr-processing} illustrates the pre-processing process. We preprocess the data only if the images are not cropped into ROIs. Given the annotation JSON with bounding boxes and corresponding text labels, we acquire the cropped images and save the processed (image, text) pairs.

For train-test splits, we used existing splits for the public datasets and generated a random split if the dataset did not provide one.

\subsection{AIHub Data License Details}

\textcolor{red}{Disclaimer: the authors are not affiliated with AIHub or with any data from AIHub.}

The data from AI Hub has been released for open public uses, including but not limited to commercial/non-commercial purposes in the research and development of AI. In order to control the data usage, downloading the data from AIHub requires an account. For further information, please refer to their \href{https://www.aihub.or.kr/intrcn/guid/usagepolicy.do?currMenu=151&topMenu=105}{policy page}.

\end{document}